\documentclass[review]{elsarticle}
\pdfoutput=1

\usepackage{lineno,hyperref}
\usepackage{times}
\usepackage{epsfig}
\usepackage{graphicx}
\usepackage{amsmath}
\usepackage{amssymb}
\usepackage{multirow}
\usepackage{rotating}
\usepackage{booktabs}
\usepackage{float}
\usepackage{caption}
\usepackage{subcaption}
\usepackage{verbatim}

\usepackage{cleveref}
\crefname{section}{§}{§§}
\Crefname{section}{§}{§§}

\usepackage{amssymb}

\journal{Computer Vision and Image Understanding}

\begin{document}

\begin{frontmatter}



\title{Automatic Action Annotation in Weakly Labeled Videos}

\author{ Waqas Sultani, Mubarak Shah}
 \address{Center for Research in Computer Vision, University of Central Florida, USA\\ waqassultani@knights.ucf.edu, shah@crcv.ucf.edu}

\begin{abstract}
Manual spatio-temporal annotation of human action in videos is laborious, requires several annotators and contains human biases. In this paper, we present a weakly supervised approach to automatically
obtain spatio-temporal annotations of an
actor in action videos. We first obtain a large number of
action proposals in each video. To capture a few most representative
action proposals in each video and evade processing
thousands of them, we rank them using optical flow and saliency in a 3D-MRF based
framework and select a few proposals using MAP based proposal subset selection method. We demonstrate that this ranking preserves the
high quality action proposals. Several such proposals are
generated for each video of the same action. Our next challenge
is to iteratively select one proposal from each video
so that all proposals are globally consistent. We formulate
this as Generalized Maximum Clique Graph problem using
shape, global and fine grained similarity of proposals
across the videos. The output of our method is the most
action representative proposals from each video. Our method can also annotate multiple instances of the same action in a video. We have
validated our approach on three challenging action datasets: UCF Sport, sub-JHMDB and THUMOS'13 and have obtained promising results compared to several baseline methods.  Moreover, on UCF Sports, we demonstrate
that action classifiers trained on these automatically obtained
spatio-temporal annotations have comparable performance to the
classifiers trained on ground truth annotation.
\end{abstract}


\end{frontmatter}


\section{Introduction}

Despite the recent advances in computer vision, action recognition and detection in realistic videos is a problem far from being solved. In action recognition, the main goal is to classify whether the testing video clip contains a specific action or not irrespective of the location of an actor. In addition to recognition in action  detection, we also want to know the precise spatio-temporal location of an actor. Both problems have their own challenges, with the latter being harder than the former.

Action detection methods accuracy lag far behind action recognition algorithms \cite{Yicong-cvpr2013,Wang14,Mihir}, which perform quite well on extremely challenging datasets \cite{IDTF,THUMOS13,IIDTF}. One of the reasons is that, in addition to the difficulty of problem, there are only few detection datasets available, which limits the opportunity to train, validate and test new methods. This is because action detection datasets require precise spatio-temporal bounding box annotations for each video, where the spatio-temporal annotations are cumbersome to obtain, require many human annotators, hundreds of hours, expensive annotation interfaces and are subject to human biases. Moreover, for any new action class, the annotation needs to be done from scratch.
As action datasets are exponentially growing, design and development of generic automatic annotation methods are very much needed. This does not only reduce human biases, but also saves time and cost.

With the advent of large image and object datasets \cite{Everingham15,ILSVRC15}, automatic object annotation is  becoming challenging and hence, gaining more attention from the research community. To address this issue, several approaches have been presented recently to obtain object level annotations from image level annotations. These approaches attempt to automatically obtain object bounding box location using: eye-tracking \cite{EyeTracking}, transferring annotations from previously annotated object to the new class \cite{KnowledgeTransfer}, exploiting generic object knowledge \cite{GenericKnowledge}, jointly localizing objects in multiple images \cite{Col-14-kevin,Prest} and using video \cite{TangECCV14}. The straightforward extension of these approaches to automatically obtain action spatio-temporal annotations from video level labels is not feasible because temporal domain is quite different from spatial domain \cite{Yicong-cvpr2013}. Temporal length of an action can be arbitrarily long depending upon action cycles captured in a video. In addition, 3D cuboids would contain significant background pixels due to large camera motion and spatial motion of an actor.

Instead of 3D cuboids, spatio-temporal action proposals obtained through segmentation \cite{Mihir,GBH-ECCV2012} or dense trajectories \cite{IIDTF, APT} of a video can precisely enclose action boundaries and capture arbitrary spatio-temporal action localization.  We believe that the action proposals can provide a useful platform to obtain automatic action annotations in videos. However, these methods produce a humongous number of action proposals; some precisely enclose the complete action, while the majority are noisy and capture only action parts, background, camera motion, or both foregrounds-backgrounds. Our goal is to automatically discover the most action representative proposals from each action video that tightly covers actor spatio-temporal localization. We propose to obtain these spatio-temporal action annotations using videos level.

\begin{figure}[t]

\begin{center}
  \includegraphics[width=12.5cm,height=5.5cm]{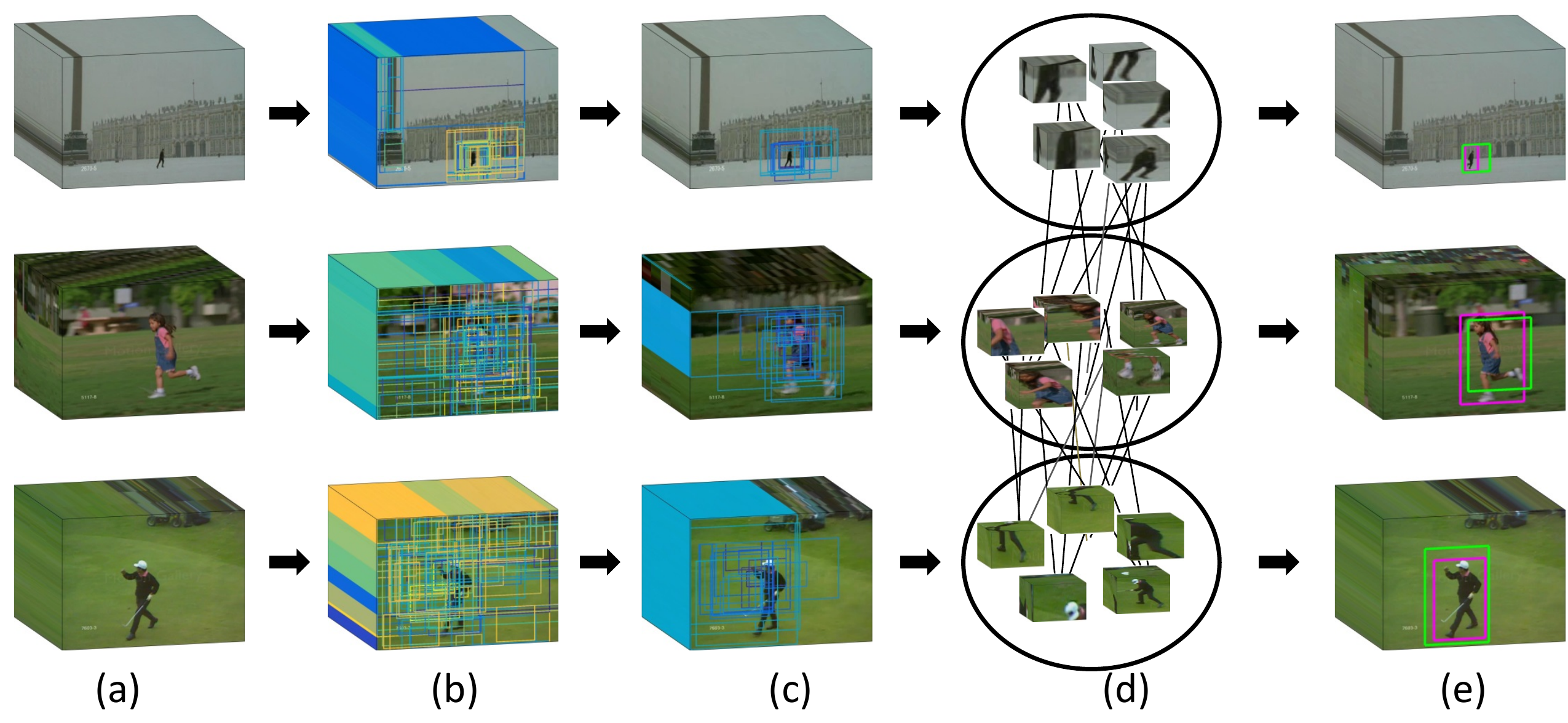}
\end{center}
\caption{Block diagram of our approach. (a) Given the multiple videos of the same action ('running' in this figure), (b) We first compute large number of action proposals in each video ($\S 3.1)$, (c) After that we obtain a few most action representative proposals in each video using motion and saliency information employing MAP based proposal subset process ($\S 3.2)$, (d) Then, we construct a fully connected graph between proposals across multiple videos, where edge between proposals captures global, fine grain and shape similarities between proposals ($\S 3.3)$, (e) Finally, using generalized maximum clique of this graph , we obtain the most action representative proposal in each video (($\S 3.4)$).  Colors of proposals are randomly selected except (e) where magenta shows ground truth and green box represents automatically discovered action proposal (actual results).}\label{fig:BlockDiagram}\vspace{-0.2in}
\end{figure}

In this paper, we present a simple yet effective weakly supervised approach to obtain the bounding box annotation of an action. The block diagram of our approach is shown in Figure \ref{fig:BlockDiagram}.  Given action proposals, we seek to discover automatically the proposals that have the higher probability of representing spatio-temporal location of an actor. Given a large number of proposals, we initially rank them according to their probabilities of being representative of an action. We achieve this using MAP based subset selection procedure by employing optical flow gradients and saliency in the 3-D MRF based framework.
We then utilize similarity between top ranked proposals across different videos of the same action and re-rank the proposals. For this purpose, 
we build a fully connected graph where all proposals in one video are connected to every proposal in all other videos and the edges between proposals capture global, fine grain and shape simiarities between proposals. Finally, we formulate the proposals matching across multiple videos as a Generalized Maximum Clique Problem (GMCP) \cite{Amir-pami2014,GMCP-2003}. The output of our method is  the most action representative spatio-temporal locations in the video.

Our method is weakly supervised, since we only use video level labels instead of bounding box level annotations. It is efficient since we achieve the final bounding box action annotation within a few seconds using GMCP employing only a few top ranked action proposals. Our approach is useful, as it can seamlessly be integrated with any other action detection method. With these key aspects, our method satisfies three main characteristics of a visual system: less supervision, efficiency and usefulness.

The organization of the rest of the paper is as follows: in section 2, we review related work on weakly supervised object and action localization. In section 3, we describe proposed approach in detail. We report results in section 4 and section 5 concludes the paper.

\section{Related Work}\label{sec:foo}

Recently, there has been growing interest in solving the challenging problem of human action localization \cite{Yicong-cvpr2013,PatternSearch,FastActionDetection}.  Most of these approaches are inspired by object localizations, extending the detection problem from 2D bounding box to 3D cuboid.  Due to a fixed size of the cuboid, detection results of these methods contain a significant portion of background, specifically, when actors aspect ratio varies significantly.  To circumvent this problem, more precise action detection approaches have been introduced \cite{Mihir,Wang14,APT}.  However, training action classifiers using any of the above methods requires hundreds of time consuming spatio-temporal annotations.

In order to avoid these time-consuming annotations, weakly supervised methods have been introduced recently for training action classifiers \cite{weaksimon-eecv12,weak-BMVC-siva,Zain-BMVC2014}. Boyraz et al. \cite{Zain-BMVC2014} presented a weakly supervised action recognition method to estimate discriminative regions in each frame. The histograms of these discriminative regions are used for learning the action classifier using two-layer neural networks. Similarly, \cite{weaksimon-eecv12} reported a method where discriminative regions are considered as latent variables. They proposed similarity constrained latent SVM, which jointly learns the action classifier as well as discovers discriminative regions. Both of these methods have shown improved classification accuracy using discriminative regions without requiring manual spatio temporal annotations. Although they provide improved results, as mentioned in the papers \cite{weaksimon-eecv12, Zain-BMVC2014}, automatically discovered discriminative regions do not necessarily represent human action locations. The authors in \cite{weak-BMVC-siva} proposed a weakly supervised action detection method based on multiple instance learning. However, one of the major limitations of their method is the assumption that actions can only be performed by standing persons. Therefore, their method is not applicable to recent datasets \cite{action-mach-08,THUMOS13, Kuehne11} which contain huge articulated human motion.

Recently, there has been a lot of interest in weakly supervised object localization using multiple images and videos \cite{TangCVPR14}\cite{TangECCV14}. These approaches compute object candidate locations using the objectness score \cite{Objectness} and find the similar boxes in multiple images or videos frames to improve object localization.  To the best of our knowledge, no such analysis has been presented for action localization, before.

A related problem includes video object co-segmentation \cite{Dong-ECCV14,Co-seg-CVPR14}. Instead of using each video independently, these methods use multiple videos to improve segmentation of the moving objects. The authors in \cite{Dong-ECCV14} obtain an object proposal in each frame and track them over the video, forward and backward. Final segmentation is achieved using shape, color and motion similarity in a regulated maximum weight clique's framework. Similarly, \cite{Co-seg-CVPR14} produced accurate co-segmentation of a moving object in a video using shape and color similarities employing CRF. Both of these methods have been tested on clean videos where we test the proposed approach on challenging action datasets that have large variations in pose, cluttered backgrounds, poor illumination conditions and low quality videos.

\section{Proposed Approach}

In the proposed approach, we begin with obtaining action proposals in a video. In each video, we rank action proposals using elementary action cues and select a few high quality action proposals from several thousand proposals. Then, for  multiple videos of the same action, we compute similarity between proposals across videos by carefully considering their saliency, shape, and fine grain similarity. Finally, by using the similarity information among multiple proposals in a global framework, we select the most action representative proposal in each video.

\subsection{Action Proposals}

\begin{figure}[b]
\begin{center}
   \includegraphics[width=13cm,height=3.cm]{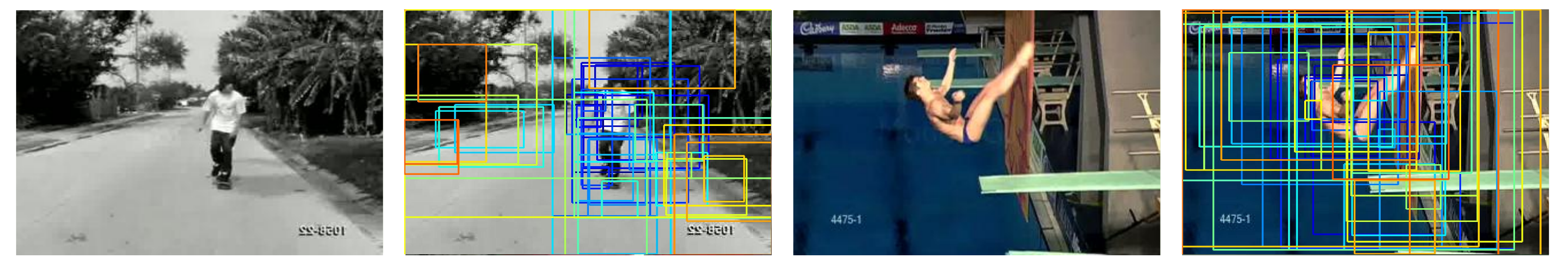}
\end{center}
 \caption{Typical action proposals. Color of proposals is randomly assigned.}
\label{fig:AllProposals}

\end{figure}

The first step of our approach is to generate spatio-temporal action proposals. To achieve this, we  obtained action proposals using improved dense trajectories\cite{IIDTF} using an unsupervised method \cite{APT}. Similar to \cite{APT}, we employed unsupervised hierarchical clustering algorithm \cite{SLINK}  to merge dense trajectories using HOG, HOF, MBH, Traj  and SPAT(spatio-temporal positions) features. To achieve efficiency and spatio-temporal smoothness, only few nearest neighbors are considered to compute similarity, while merging clusters. Finally, the clusters whose distance is more than  a certain threshold represents individual action proposals.   As compared to previous methods \cite{Danoneata,Mihir}, this method \cite{APT} does not require supervoxel segmentation and has time and space complexity of $O(n^2)$. Figure  \ref{fig:AllProposals} shows typical action proposals for UCF-Sports dataset.

We stress that our main approach of action annotation  does not depends on any specific  action proposal method and any recent action proposal methods \cite{Mihir,Danoneata,APT} can be employed .

 \begin{figure}[b]
\begin{center}
   \includegraphics[width=12cm,height=4.46cm]{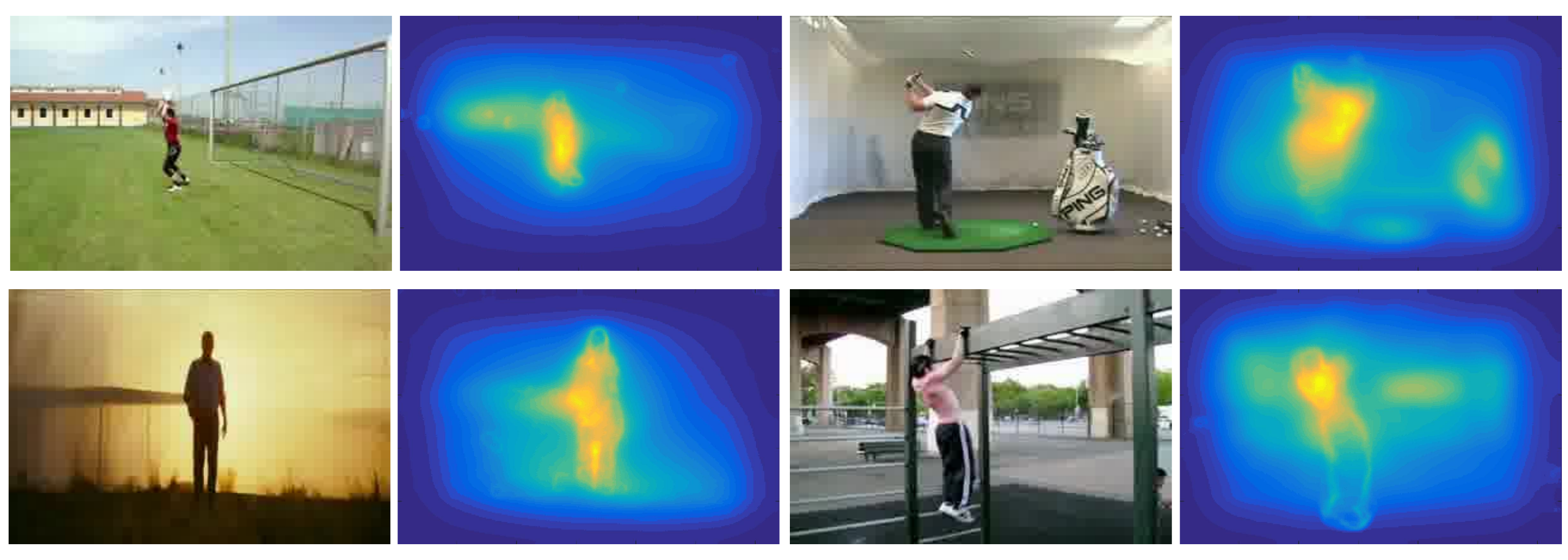}
\end{center}
 \caption{Action Score Map, $\Lambda_s$, for four actions videos of Sub-JHMDB dataset.}
\label{fig:MRF_Pic}

\end{figure}

\subsection{Initial Proposals Ranking}
Although action proposals reduce search space for action detection and classification, the number of proposals are still huge and cannot be  directly used in place of  action annotation. However,  given the large number of proposals, one can safely expect at least a few proposals that would have very high overlap with the actual actor spatio-temporal location. Our ultimate goal is to discover those action representative proposals automatically.

Not all action proposals are equally important. Many proposals originate from the background and several contain only part of an action. Moreover, computing expensive features from all proposals is computationally inefficient. Therefore, we propose to rank action proposals using simple elementary features and keep  a few highly action representative proposals, only. Inspired by  \cite{C14,Objectness}, we compute the following action cues from each video independently.

\noindent\textbf{Motion Cues:}
Motion boundaries have proven to be resistant to camera and background motion but characterize human motion quite well \cite{IIDTF}. Therefore, we compute frobenius norm of optical flow to estimate the probable location of an actor.

\begin{eqnarray}
& \left\|U_X\right\|_F =& \left\|\begin{bmatrix} u_{x} & u_{y}  \\v_{x} & v_{y} \end{bmatrix}\right\|_F \phantom{i}
\vspace{-0.1in}
\end{eqnarray} where $U_X$ represents forward optical flow , $u_x$, $v_x$, $u_y$ and $v_y$ are optical flow gradients.



%

\noindent\textbf{Visual saliency:}
Actors usually stand out among their neighbor and capture visual attention. We estimate saliency of each pixel in video frame using \cite{GraphBasedVisualSaliency}.  In this method, feature and orientation maps are computed at multiple scales using local gradients and Gabor filters, respectively. Finally, center surround activation maps and their normalization are obtained using a fully connected graph over feature space. Further details of this method can be found in \cite{GraphBasedVisualSaliency}.

\noindent\textbf{Spatio-temporal coherence:}
The above features are estimated independently for each frame and max normalized to represent foreground score maps.  We aggregate
motion (M) and saliency (S) as: M+S. These initial scores have no or little spatial temporal coherence.  Therefore, we impose spatio-temporal consistency using a discontinuity preserving 3D Markov Random Field framework, which enforces smoothness in nearby video locations. The video is considered as a 3D grid graph, where each node (pixel) is connected to four spatial neighbors and two temporal neighbors. . Formally, 3D MRF energy minimization is given as,
\begin{equation}
E(l)= \displaystyle\sum\limits_{p\in \mathcal{V}} \Phi(l_p) + \displaystyle\sum\limits_{(p,q)\in
\mathcal{N}} \Psi(\l_p - \l_q),\vspace{-0.1in}
\end{equation}
where $l_p$ is labelling (score) of pixel \emph{p}. We used quadratic unary term $\Phi$ and truncated quadratic smoothness term $\Psi$.
The inference over this graph is achieved using Max-Product/Min-Sum loop belief propagation \cite{Pedro-Seg,C14}. Qualitative examples of foreground scores, in Figure \ref{fig:MRF_Pic}, for moving camera and low quality videos demonstrate the robustness of the above framework for estimating foreground regions. 

Finally, we estimate initial action score, $\Omega_{p_i}$, of each proposal by computing  foreground score (normalized by proposal area) within each proposal.

Given all proposals $\mathcal{P}$ in a video $\mathcal{V}$, we use this initial action score to select a few most probable action proposals. However, this initial action score of proposal can be noisy and also there are many highly overlapped proposals, therefore to select a small subset $\mathcal{S}$ of most probable action proposals, we propose to use  MAP-based proposal subset selection procedure similar to \cite{NewNMS}.   

Overall, we want to group proposals into automatically determined number of clusters and select exemplar proposal from each cluster. We assume an auxiliary variable $\textbf{Z}$= $(z_i)^n_i$, where $z_i$ =$j$ if proposal $p_i$ belongs to a cluster represented by proposal $p_j$.  The joint distribution of a few selected proposals $\mathcal{S}$ and  $\textbf{Z}$ for video $\mathcal{V}$ is given by:
\begin{equation}
P(\mathcal{S},\textbf{Z}|\mathcal{V})=\frac{P(\mathcal{V}|\mathcal{S},\textbf{Z})P(\mathcal{S},\textbf{Z})}{P(\mathcal{V})},
\end{equation}
where $P(\mathcal{V}|\mathcal{S},\textbf{Z})$ represents likelihood term and $P(\mathcal{S},\textbf{Z})$ represents prior term. 
We want to estimate Maximum a posteriori (MAP) of above equation.  In order to select a few action representative less overlapping proposals, the prior term can be written as \begin{equation}
P(\mathcal{S},\textbf{Z})=K_1P(\textbf{Z}))\mathcal{W(\mathcal{S}) }\mathcal{C}(\mathcal{S},\textbf{Z}),
\end{equation}
where $K_1$ is a normalization constant.  $\mathcal{C}(\mathcal{S},\textbf{Z})$ is 1 for exemplar proposal from each cluster.
 $\mathcal{W(S)}$ is prior information about detection window and is given as:
\begin{equation}
\mathcal{W}=W_1 \times W_2,
\end{equation}
 $W_1$ softly penalize highly overlapped proposal windows in $\mathcal{S}$ and is given by:
\begin{equation}
W_1= \prod_{i,j:i \neq j} exp(-\gamma \times  IOU(p_i,p_j)),
\end{equation}
where $IOU(p_i,p_j)$ represents intersection over union between two proposals.
$W_2=exp(-\phi N)$ controls the number of finally selected proposals. We choose this parameter so that at least one hundred proposals in each video are produced.

After substituting prior and likelihood term, Equation 3 can be written as:
\begin{equation}
P(\mathcal{S},\textbf{Z}|\mathcal{V}) \propto P(\textbf{Z}|\mathcal{V}) \mathcal{W(\mathcal{S}) }\mathcal{C}(\mathcal{S},\textbf{Z}),
\end{equation} where $\mathcal{W(\mathcal{S}) }$ and $\mathcal{C}(\mathcal{S},\textbf{Z})$ are defined above and considering the independent assumption among $z_i$,  $P(\textbf{Z}|\mathcal{V})$  can be given as follows:
\begin{equation}
P(\textbf{Z}|\mathcal{V})= \prod_{i=1}^n P(z_i|\mathcal{V}),
\end{equation}
In the above equation, $P(z_i=j|\mathcal{V})=\lambda \times K_2 $ if $j=0$, otherwise it is $K_2 \times IOU(p_i,p_j)s_i$, where$K_2$ is a normalization constant. Note that  $P(z_i=j|\mathcal{V})$ encourage proposals which have high overlap with many high initial action score proposals and hence it is robust to individual noisy score proposal.

\begin{figure}[t]
\begin{center}
   \includegraphics[width=13cm,height=3.cm]{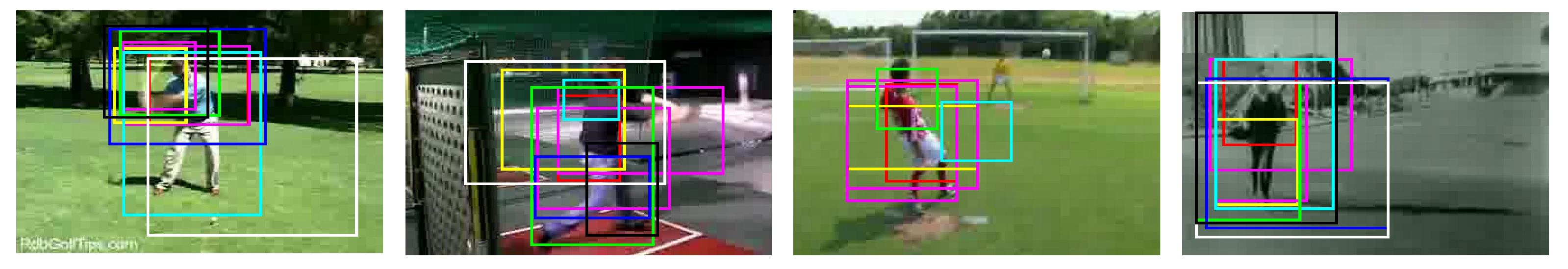}
\end{center}
 \caption{Top few action proposals for four actions videos of Sub-JHMDB dataset.}
\label{fig:AfewTopProposals}

\end{figure}

Figure \ref{fig:AfewTopProposals}, shows a few top ranked action proposals after proposal subset selection process.

\begin{figure}[t]
\makebox[\linewidth][c]{
\begin{tabular}{c@{\hspace{0.01in}}}
 \epsfig{file=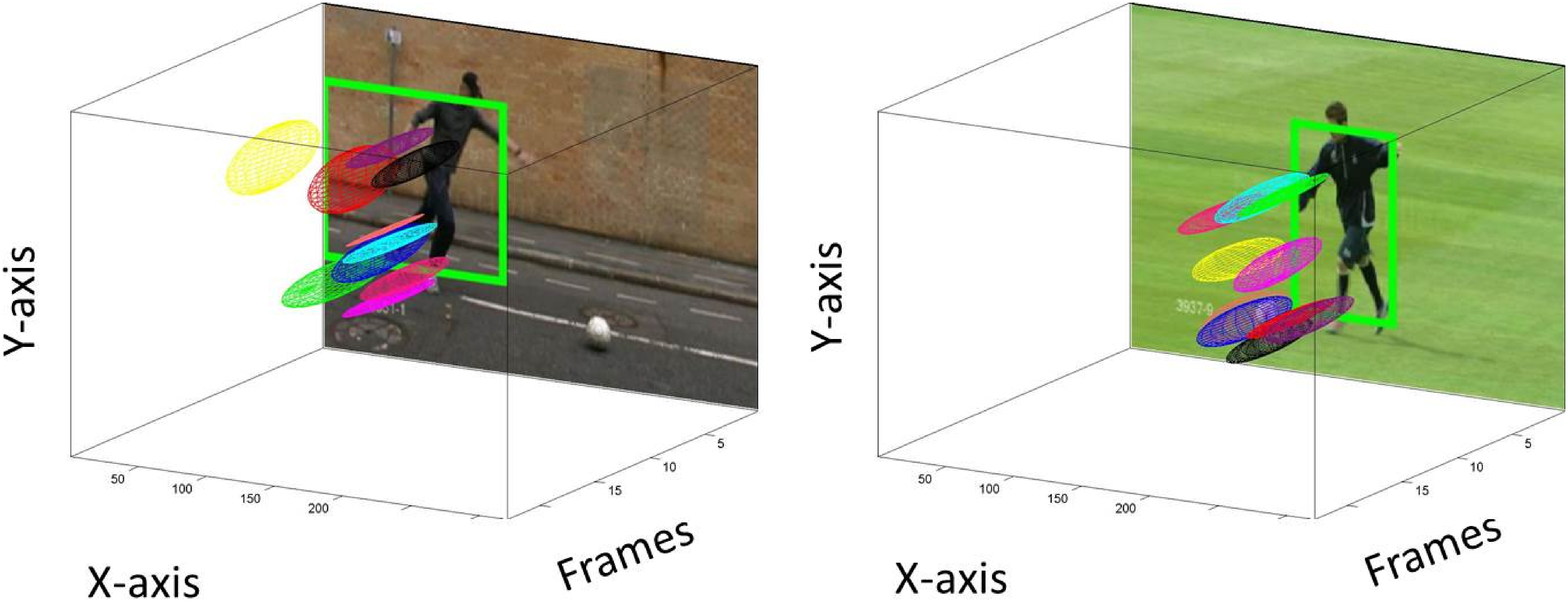,width=0.95\linewidth,height=1.7in,clip=}\vspace{-0.01in}\\
 \epsfig{file=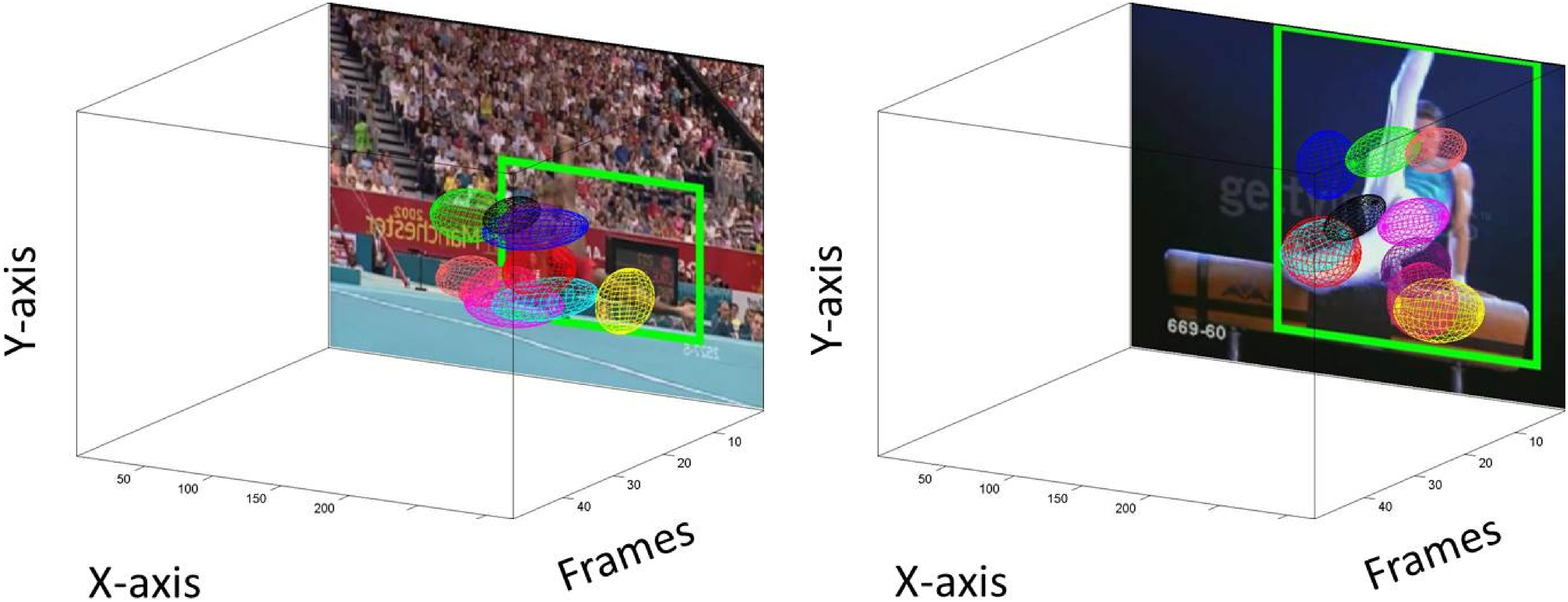,width=0.95\linewidth,height=1.7in,clip=}\vspace{-0.01in}\\



%
\end{tabular}}

\caption{Illustration of proposals matching across videos. The similar color ellipses show the matching clusters using Hungarian algorithm.}\label{fig:Clusters}\vspace{-0.2in}
\end{figure}

\subsection{Proposals Similarity Across Multiple Videos}

Although a few top ranked proposals maintain high MABO (defined in Equation 12), the top most  proposal does not necessarily represent the best available proposal. Therefore, we re-rank the proposals by leveraging action proposals similarity across multiple videos of the same action.

A naive similarity measure between proposals can  hurt the proposal ranking, since, sport videos backgrounds are more similar than the action itself. Therefore, we use global, fine grain and shape similarities between proposals to discover the most action representative proposals. 
 
Each of the similarity measure between proposals is explained below.


\noindent\textbf{Global Similarity}

We use bag of words (BOW) similarity between proposals.  
We represent each proposal by $M$-bin global histogram and spatial pyramids of $2\times 2$ using improved dense trajectory features (Trajectory, MBH, HOF and HOG) \cite{IIDTF}.
Next, the similarity between two proposals is measured using $\chi^2-$ distance, which is defined as:
\begin{eqnarray}
S_{ij}^f =exp \left(-\gamma \sum_{k=1}^{k=d}\frac{(h_{ik}-h_{jk})^2}{(h_{ik}+h_{jk})} \right),
\end{eqnarray}where $h_i$ and $h_j$ respectively, represent bag of words histogram of feature $f$ for $i^{th}$ and $j^{th}$ proposal and $d $ is the dimensions of the histogram. The final similarity between any two proposals, $\bf \Theta_{ij}$, is the linear combination of individual feature similarities.

\noindent\textbf{Fine Grain Similarity}

Proposal matching using spatial pyramid (the fixed grid structure) has an underlying assumption that similar action parts appear at the same location in both proposals. However, due to actor articulations, large camera motion, pose and scale variations, the fixed location assumption is not always true. Therefore, we propose the use of flexible matching between action regions to obtain aggregated similarity between action regions as a proposals similarity measure. Since the flexible similarity measure takes into account the similarity across the local region between proposals, we call it fine grain similarity measure.

To achieve this, we cluster \emph{raw} improved dense trajectory features within each proposal in $C_r$ clusters, where subscript \emph{r} corresponds to MBH, HOG, HOF or Trajectory. A $C_r \times C_r$ distance matrix is computed using Euclidean distance. To allow the flexibility in matching between different spatio-temporal action regions, we use only raw features (without their actual coordinates) during Euclidean distance computation. Finally, the optimal one-to-one matching between clusters across two proposals is obtained using Hungarian algorithm \cite{Kuhn1955}.  We compute similarity between clusters of each raw feature separately and final similarity, $\bf \Gamma_{ij}$, is the linear combination of all of them. Figure \ref{fig:Clusters} illustrates the flexible matching of clusters (the same color) across proposals. In experiments we use insensitive parameter $C_r$=6.\\

\noindent\textbf{Proposal Shape Similarity}

In addition to spatio-temporal features within action proposals, the shape of proposal windows (height, width and aspect ratio) over time itself carries useful information about an action. Mostly the same action in multiple videos undergo through similar articulations and  therefore, the similar shape proposal windows across videos likely capture  the same action. 

We define the shape of action proposal $p_x$ and $p_y$ over time as:

\begin{equation}
\begin{split}
\Lambda_{p_x}=[r_1,r_2,\ldots,r_n] \\
\Lambda_{p_y}=[r_1,r_2,\ldots,r_m]
\end{split}
\end{equation}
where $r_i=\frac{w_i}{h_i}$ and  $w_i$ and $h_i$ are the width and height of proposal in frame $i$ and $m$, $n$ are length of proposals.  A naive way to match shape of $p_x$ and $p_y$ is to match $r$ values frame by frame. However, the same action can occur with different speeds in different videos and therefore, in most cases $n \neq m$. Hence, we propose to  consider proposals' shape  $\Lambda_{p_x}$ and $\Lambda_{p_y}$ as time series and find similarity, $\bf{ \Pi_{ij}}$, between them using dynamic time warping.

\subsection{Generalized Maximum Clique Graph Optimization}

Given each proposal action score, as well as their pairwise similarity across multiple videos of the same action, we seek to identify the most action representative proposals from every video that  have a high action score as well as a high  similarity with highly action representative proposals in other  videos.  Due to large intra-class variation, matching only two videos may not necessarily facilitate better localization. Therefore, we re-rank all the videos jointly using a global framework.

To this end, a fully connected graph $ \textbf{Z = (V, E)}$ is constructed, such that $\textbf{V} = \{v_i\}$, $i \in \{1,\ldots,n\}$, is
the set of all proposals, and $ \textbf{E} = \{e_{ij}\}$, $i \in \{1,\ldots,n\}$, $j \in \{1,\ldots,n\}$, represents the edge between ${p_i}$ and ${p_j}$, where $e_{ij}= \eta_{pi}\times \eta_{pj}(\bf \Theta_{ij}+\Gamma_{ij}+\Pi_{ij})$.  $\eta_{pi}$ discourage proposals whose length is very small as compared to the length of video and is given by: $exp(-(m-n)/n)$, where $m$ is length of video and $n$ is length of proposal. 

We divide all nodes (which correspond to proposals) into disjoint groups, where each group $Z_i$ belongs to one action video. The nodes within each group is a set of all top ranked proposals in a single video. We call them a group because they belong to the same video. Since we want to select one node from each group, the feasible solution is a subgraph that satisfies two constraints: 1) Only one node from each group is selected; 2) If one node is included in feasible solution, then its $N-1$ edges to single node in each of $N-1$ groups should be included as well.

Formally, the feasible solution can be found by maximizing the following objective function:
\begin{equation}
  \sum_{i=1}^{i=N} \sum_{j=1,j\neq i}^{j=N}   \Bigg( \alpha \Omega_{p_i}+ (\bf{\Theta_{ij}}+\bf{\Gamma_{ij}}+\bf{\Pi_{ij}})\eta_{pi}\times \eta_{pj}\Bigg),
\vspace{-0.05in}
\end{equation}
where $\Omega_{p_i}$ represents the initial action score of  $i^{th}$ proposal, $\bf \Theta_{ij}$, $\bf \Gamma_{ij}$ and $\bf \Pi_{ij}$ show the similarity between proposals computed in previous section, $\eta_{pi}$ discourage small length proposals and $\alpha$ controls the weight of the initial action score for the final objective function.

 The optimal solution is a subgraph that maximizes the above objective function.  It is easy to observe that the above combinatorial optimization problem falls under the umbrella of generalized maximum clique problem (GMCP) \cite{Amir-pami2014}\cite{GMCP-2003}. GMCP is the class of graph theory problems that generalizes the standard subgraph problems (from node to group of nodes). The input to GMCP is graph $\textbf{Z}$, as defined above. Specifically, the input graph consists of groups of nodes where  edges exist between all the proposals across the groups only with no connection within the group. The output of GMCP is a subgraph $ \textbf Y_s$ $= (V_s, E_s)$, such that each node in the subgraph belongs to one video only and the objective function is maximized.

 GMCP is an NP-hard problem.
We have used the approximate solver proposed in \cite{Amir-pami2014}, for which code is available online. This local neighborhood solver has fast convergence speed and is memory efficient. Specifically, we initialized the initial solution from the top ranked proposals (from section 3.2) and generated $N_Z$ $\times$ $Z_N$ local feasible solutions of size 1, where $N_Z$ denotes the total number of groups (number of videos of an action) and $Z_N$ represents the number of nodes in a group (number of proposals in a single video). The solution that has the maximum score is selected and again $N_Z$ $\times $ $Z_N$ local feasible solutions are generated around this newly found solution and so on. We repeat this process until we reach the maximum number of iterations or no more updated solution can be obtained with further iterations.

The above formulation of GMCP assumes only one action instance in a  video. However, there are some videos in UCF-Sports and THUMOS'13 datasets which have multiple instances of an action in the same video. To annotate multiple action instances in a video, we use GMCP iteratively.  During each iteration (after the first one), we stop the node selection for the videos that have all of its instances annotated. For the videos which have yet more instances to be annotated, we ignore the nodes that have high overlap with already selected node (as they may be localizing the same instance), and find GMCP solution from rest of the nodes.  We repeat this process until all instances in all videos are annotated.

\section{Experimental Results}

We have evaluated our method on three action datasets: UCF Sports \cite{action-mach-08}, sub-JHMDB \cite{Kuehne11,Jhuang:ICCV:2013} and THUMOS13 \cite{THUMOS13}. These datasets are among the most challenging action datasets.   Ground truth bounding box annotations are available for all three datasets.

In all experiments, we compute action proposals using online implementation of  \cite{APT}. Improved dense trajectory features are extracted using \cite{IIDTF} and encoded in standard bag of words paradigm. The value of $\alpha$ in Equation 11, which controls the contribution of initial raking, is set to 0.07.

\textbf{UCF Sports} \cite{action-mach-08}  contains 10 human actions.  This dataset includes actions such as:  diving, kicking, lifting, horse riding, etc. These low quality YouTube videos contain huge camera motion, dynamic backgrounds, view-point changes and large intra-class variations.

\textbf{sub-JHMDB} \cite{Jhuang:ICCV:2013} contains 12 complex human actions.  This dataset includes actions such as:  catch, climb stairs, run, jump, swing basketball etc. This dataset is a subset of JHMDB \cite{Jhuang:ICCV:2013} and contains 316 videos. As mentioned in \cite{Jhuang:ICCV:2013}, this subset is far more challenging as compared to the whole JHMDB dataset.

\textbf{THUMOS13} \cite{THUMOS13} is the largest and the most challenging trimmed action detection dataset with 24 complex human actions. It includes actions such as:  pole vault, skiing, ski-jet, surfing, fencing, cricket bowling etc. This dataset is a subset of UCF-101 and includes 3207 videos with multiple instances of an action in the same video.

First, we evaluate initial proposal ranking followed by qualitative and quantitative analysis of localization results and their detailed analysis.
\begin{table}[b]
\begin{center}
\scriptsize{
\begin{tabular}{l@{\hspace{0.1in}}c@{\hspace{0.1in}}c@{\hspace{0.1in}}c@{\hspace{0.1in}}c}
\toprule[1.5pt]
Proposals  &   UCF Sports     &  Sub-JHMDB  & THUMOS13 \\
\cmidrule(r){1-4}
Top 100   &   56.01   &  55.25 &   35.26  \\
All (Upper bound)   &   62.40   &  57.77 &   46.71 \\
\toprule[1.5pt]
\end{tabular}}
\end{center}
\caption{ The first row illustrates MABO using top ranked proposals in  the video.  The bottom row shows the MABO using all proposals in a video. On average, UCF Sports, Sub-JHMDB and THUMOS13, respectively, contain 1866, 328 and 2300 proposals in every video.}
\label{table:MABO}\vspace{-0.05in}
\end{table}

\subsection{Evaluation of Initial Proposal Ranking}
Following previous works \cite{Prest,Mihir,Selective}, we evaluate robustness of our initial action proposal ranking using Mean Average Best Overlap (MABO). MABO
measures the quality of the best available proposal. To compute MABO, we first compute mean of best overlap (ABO) for each action class $c$ as follows:

\begin{eqnarray}
ABO_c =\frac{1}{|K_c|} \sum_{g_i^c \in {G_c}} \max_{p_j\in P} O(g_i^c,p_j),
\end{eqnarray} where $g_i^c$ represents ground truth annotation for $i^{th}$ video in class $c$ and $p_j$ is the  $j^{th}$ action proposal from $\textbf{P}$ proposals in a video. $|K_c|$ is the total number of ground truth in class $c$.
The overlap $O$ is computed using standard $intersection$ $over$ $union$ for each frame and averaged by the number of frames where either $g_i^c$ or $p_i$ exist. Finally, MABO is mean of ABO over all action classes.
First row in Table \ref{table:MABO} shows the MABO calculated using only top 100 proposals. 
Second row presents MABO calculated using  all action proposals for all three datasets.  It is impressive to note in Table \ref{table:MABO} that even by using 10\% proposals, sufficiently high MABO is maintained. This indicates that we have at least one good quality proposal among top ranked proposals. Note that although initial proposal ranking maintains high MABO for top 100 proposals,  the top most proposal have significantly low MABO (UCF-Sports: 18.54, sub-JHMDB: 31.25, THUMOS'13: 21.01). Therefore, to achieve better localization, we can perform matching among top ranked proposals only and can ignore the processing of several thousand proposals.



\tabcolsep=0.05cm
\begin{table}[b]
\begin{center}
\small{\begin{tabular}{|l|c|l|c|l|c|l|c|}
\hline
Method & UCF Sports & Sub-JHMDB & THUMOS13  \\
\hline\hline
Cosegmentation\cite{Dong-ECCV14} & 76.17 &\phantom{ii}-& - \\
Negative Mining\cite{NegativeMining} & 25.86  &\phantom{ii}87.34 & 14.39\\
CRANE\cite{TangCVPR13} & 61.18  & \phantom{ii}86.08  &14.17   \\
Ours & \textbf{85.29}&\phantom{ii}\textbf{90.51} &\textbf{41.69} \\
\hline
\end{tabular}}
\end{center}
\caption{Localization results and comparison with related work}\label{table:Comparision}
\vspace{-0.2in}
\end{table}

\subsection{Localization Results }
In this section, we describe our experimental results for weakly supervised action localization using multiple videos. At start, each video contains on average 1866, 328, and 2300 action proposals in UCF Sports,  Sub-JHMDBand THUMOS13, respectively. After initial ranking, we select the top 100 proposals from each video from UCF Sports and top 50 proposals (to reduce computation) from Sub-JHMDB and THUMOS13.

 \begin{figure}[t]
\begin{center}
   \includegraphics[width=12cm,height=4.5cm]{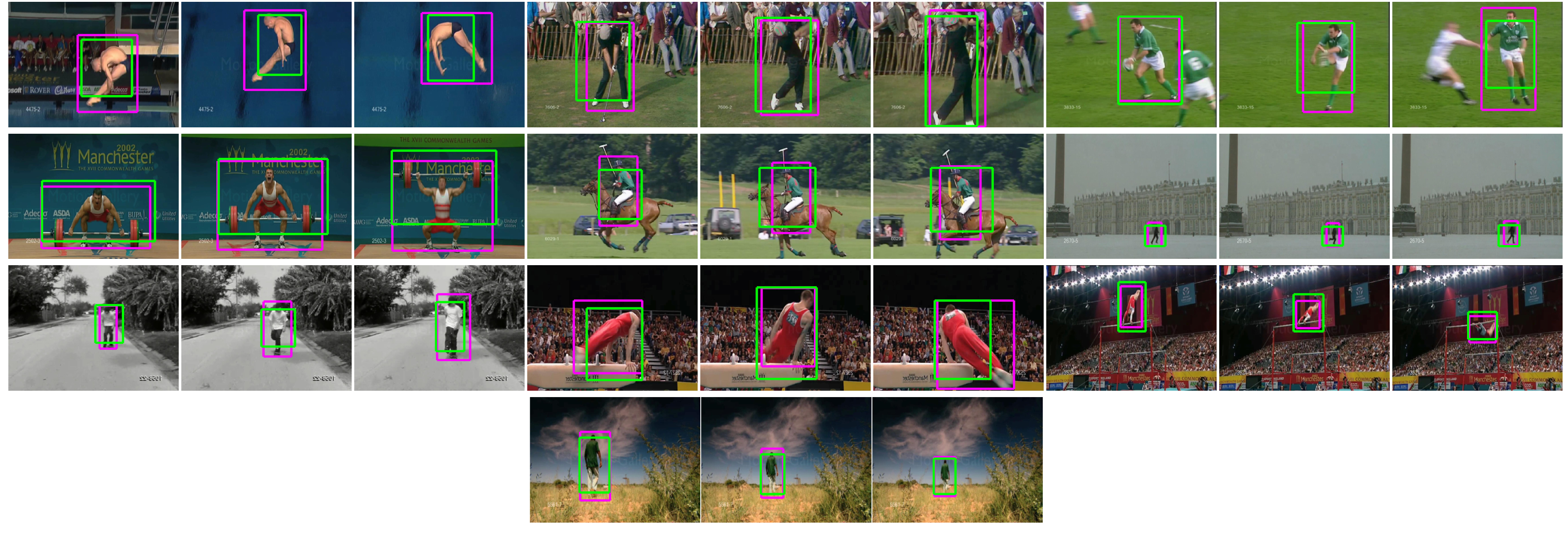}
\end{center}
 \caption{Qualitative results of all action of UCF-Sports. Every three frames show an action video at different time instances. Green and magenta bounding boxes show proposed automatic annotation and manual annotation, respectively.}
\label{fig:UCFSportsColl}

\end{figure}

Following several previous action localization methods \cite{Figurecentric2011,Yicong-cvpr2013,Mihir}, we use intersection-over-union criterion at an overlap of 20\% for correct action localization. The quantitative results for three challenging action datasets are shown in Table \ref{table:Comparision} (last row). The numbers in the table show the percentage of the videos that have correct localization. Note that we obtained these localization results without any training video.

\begin{figure}[t]
\begin{center}
   \includegraphics[width=12.1cm,height=4.6cm]{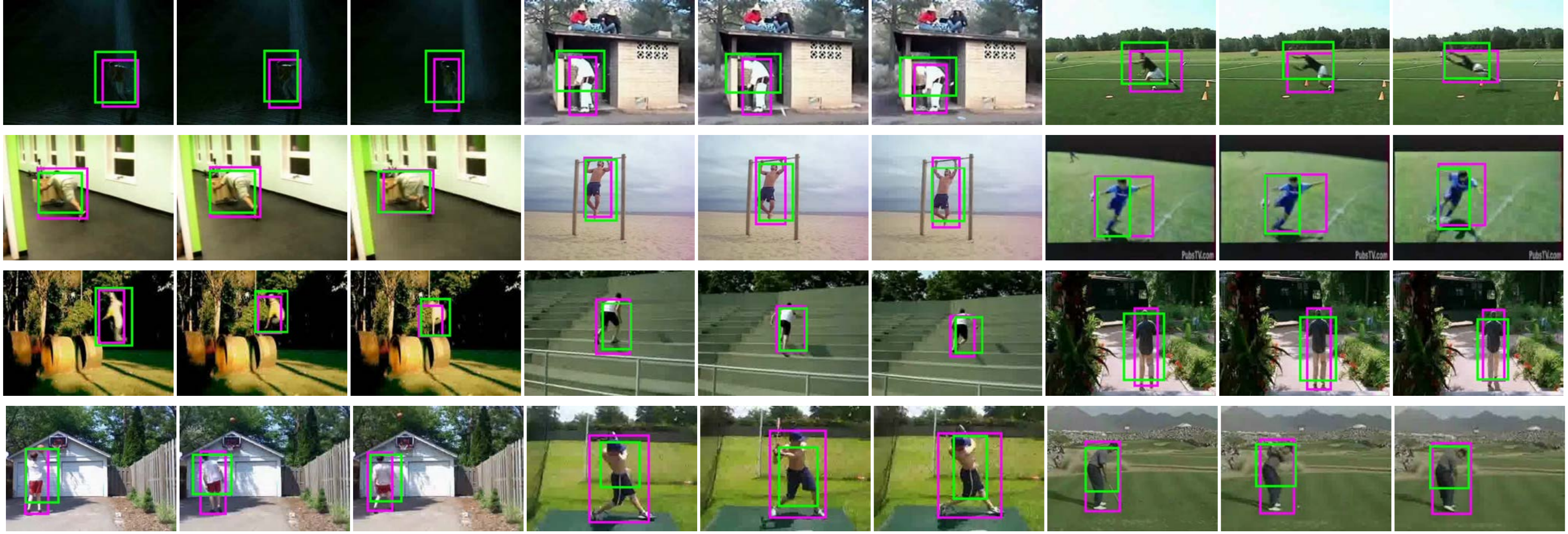}
\end{center}
 \caption{Qualitative results for all actions of sub-JHMDB. Every three frames show an action video at different time instances. Green and magenta bounding boxes show proposed automatic annotation and manual annotation, respectively.}
\label{fig:SubHMDBColl}

\end{figure}

The qualitative localization results for all action classes of UCF Sports sub-JHMDB  are shown in Figures \ref{fig:UCFSportsColl} and \ref{fig:SubHMDBColl}, respectively. 
In these figure, magenta bounding box represents ground truth annotations and green bounding box shows automatic annotation. Note that despite large camera motion and change in scale, fast and abrupt motion, background clutter occlusion, our automatic annotations closely follow the  manual annotations.

Our approach is closely related to video object co-segmentation.  For this purpose, we use a recently published video object co-segmentation method \cite{Dong-ECCV14}. Using the code available on the author's website, we produced the co-segmentation for each video and put a tight bounding box around the segmented region to represent co-localization. Experimental results of this method are given in \ref{table:Comparision}. We compared \cite{Dong-ECCV14} for UCF Sports only (because of its large time consumption: 2 min per frame).

Our work is also related to weakly supervised concept or action detection using negative mining \cite{TangCVPR13,NegativeMining}. Similar to ours, both methods assume availability of video level labels. In addition, they use negative data to localize the main concept or action in video. We follow the procedure described in \cite{TangCVPR13} and use videos not labeled with action of interest as negative data. We use both methods to discover the best representative proposals among top ranked proposals. Experimental results of  \cite{TangCVPR13,NegativeMining} on all three datasets are given in \ref{table:Comparision}. The quantitative comparison in Table \ref{table:Comparision} demonstrates the superiority of our approach.



%

\subsection{Action Detection}
\begin{figure}[h]
\begin{center}
   \includegraphics[width=6.5cm,height=5cm]{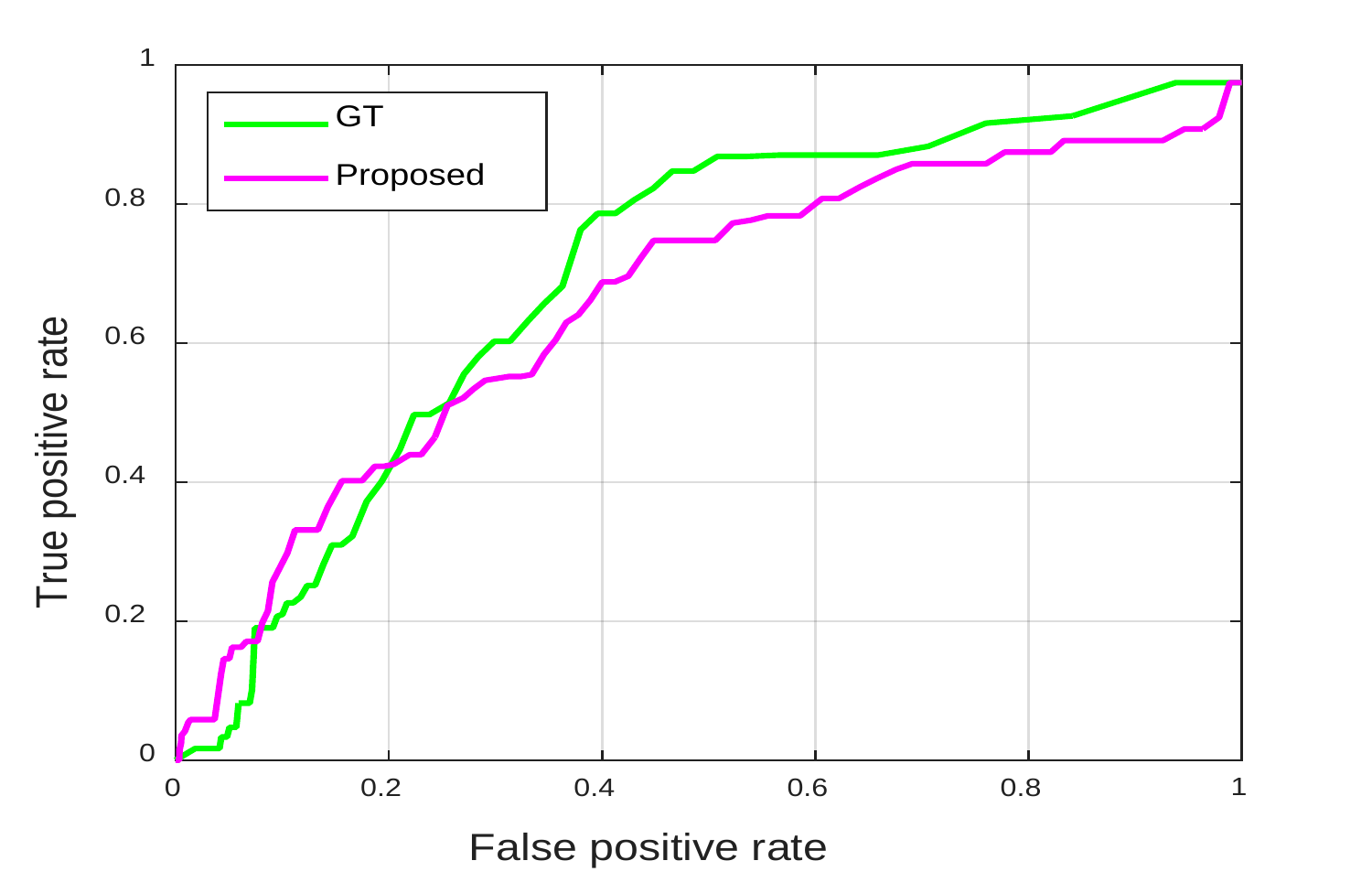}
\end{center}
\caption{The ROC curves on UCF Sports dataset. The results show comparable performance of classifier trained on manual annotations (green) and proposed automatic annotations (magenta).}\label{fig:ROCs}\vspace{-0.2in}
\end{figure}

In this section, we demonstrate the usefulness of automatically obtained bounding boxes by using them to train  action detectors  in the standard train-test settings \cite{Yicong-cvpr2013,Figurecentric2011}. To this end, we automatically annotate UCF sports actions using $\textit{training videos}$ only; $\textit{no}$ test video is used during annotation.  We use BOW representation of improved dense trajectories to  train each action classifier. Moreover, we train separate classifiers using exactly the same settings on $\textit{ground truth}$ bounding boxes. We evaluate testing results using standard metrics as shown in Figure \ref{fig:ROCs}. We consider these results quite  promising as action classifiers trained on automatically obtained annotations have comparable performance to that of classifiers trained on manual annotations.

\subsection{Analysis and Discussion}

Our approach has several components. We quantitatively evaluated each component and have shown the  experimental results for  stripped down versions of our method for UCF Sports in Table \ref{table:ComponetsContribuition}. It can be seen that each component of our method has complementary effects and helps in achieving overall localization accuracy.

\tabcolsep=0.05cm
\begin{table}[h]
\begin{center}
\small{\begin{tabular}{|l|c|l|c|l|c|l|c|l|c|}
\hline
Method &[I]nitial Score  &[I]+[S]hape & [I]+[S]+[G]lobal& [I]+[S]+[G] +Fine Grain\\
\hline\hline
Diving  & 64.2    & $\phantom{z}$64.2 & 100&$\phantom{zzzzzzz}$100\\
Golf Swing  & 5.5     &$\phantom{z}$ 55.5   & 55.5& $\phantom{zzzzzzz}$\bf{88.8}   \\
Kicking & 20.0    & $\phantom{z}$ 50.0 & 55.0 & $\phantom{zzzzzzz}$\bf{65.0} \\
Lifting & 0    & $\phantom{z}$ 0 & 50.0 & $\phantom{zzzzzzz}$\bf{100} \\
Riding Horse &  33.3    & $\phantom{z}$25.0 & 25.0 & $\phantom{zzzzzzz}$\bf{100} \\
Run  & 20.0    & $\phantom{z}$20.0 & 33.3& $\phantom{zzzzzzz}$\bf{80.0}\\
Skateboarding & 50.0    &$\phantom{z}$ 41.6   &41.6 & $\phantom{zzzzzzz}$\bf{91.6}  \\
Swing Bench & 10.0    & $\phantom{z}$15.0 &  70.0 & $\phantom{zzzzzzz}$\bf{95.0}\\
Swing SideAngle & 38.4    & $\phantom{z}$38.4 & 30.7  & $\phantom{zzzzzzz}$\bf{61.5}\\
Walk & 16.6    & $\phantom{z}$58.3 & 33.3 & $\phantom{zzzzzzz}$\bf{70.0}\\
\hline\hline
Avg & 25.8    & $\phantom{z}$36.8 & 49.4 & $\phantom{zzzzzzz}$\bf{85.2}\\

\hline
\end{tabular}}
\end{center}\vspace{-0.1in}
\caption{Component's contribution to overall performance. First column shows localization obtained using initial action score only. Second column depicts the same using proposal shape similarity as well. Third and forth column show contribuition from global and fine grain similarity, respectively.}\label{table:ComponetsContribuition}
\vspace{-0.2in}
\end{table}

Ideally, increasing the number of videos should help in getting better annotation, i.e., THUMOS'13 which has more than 100 videos per action should have better localization accuracy than UCF-Sports which has on average 15 videos per action.  However, the improvement for THUMOS'13 is less when compared with UCF Sports, mainly due to the difficulty level of THUMOS'13, as shown in the first row of Table \ref{table:MABO}. The typical behavior of localization accuracy for skateboarding action (THUMOS'13) for 100 videos is shown in Table \ref{table:Analysis}. The first row in the table shows the mean localization (mean IOU) for the batch of 25 videos. As can be seen, the videos in the 26-50 batch have  less localization accuracy as compared to other batches. Employing proposed method (second row) on the first 1-25 videos boosts their MABO  from 24.83 to 27.86. Using proposed method on 1-50 videos (third row) further increase first batch from 27.86 to 29.39 and second batch from 21.00 to 28.89. Similar pattern can be seen for third row. It is worth considering that, although localization improves in local batches (1-25, 26-50 and 51-75), the overall localization accuracy drops from 27.86 (25 videos)  to 26.55 (75 videos), mainly due to large intra-class variation in THUMOS'13 videos. 

\tabcolsep=0.05cm
\begin{table}[h]
\begin{center}
\small{\begin{tabular}{|l|c|l|c|l|c|l|c||c|}
\hline
Method & 1-25 & 26-50 & 51-75 & 76-100 & Mean\\
\hline\hline
Localization  & 24.83    &21.00& 25.65 &\phantom{i}23.75& 23.85\\
Localization(25) & 27.86    &   &  & & 27.86\\
Localization(50) & 29.39    & 28.89 &  & & 29.14\\
Localization(75) & 28.30    & 25.24 & 26.06 & & 26.55\\
Localization(100) & 28.29    & 23.57 & 26.75 &\phantom{i}26.06 & 27.42\\
\hline
\end{tabular}}
\end{center}\vspace{-0.1in}
\caption{Localization accuracy behavior across  different batches of videos. The number in brackets shows number of videos used for Localization}\label{table:Analysis}
\end{table}

\noindent\textbf{Computation Time:}
We performed experiments on desktop computer Intel Xeon E5620 at 2.4GHz. For UCF-Sports, after extracting action proposals, unoptimized MATLAB code takes 0.3$s$ for fine grain matching and 0.01$s$ for global matching between two proposals, Moreover, GMCP takes 7.7$s$ to re-rank all top proposals in all videos of an action.


\section{Conclusions}

\vspace{-0.05in}
We have presented a weakly supervised approach to automatically obtain spatio-temporal annotations in a video. In contrast to expensive and time-consuming annotations, we obtain these spatio-temporal annotation boxes in a few seconds by matching action proposals across multiple videos using their feature and shape similarities. Moreover, we have demonstrated  that these annotations can be used to learn robust action classifiers. In future work, we plan to extend our framework to localize actions without video level labels.






\bibliography{pami-actionsFromPatterns-08}

\end{document}